\documentclass[11pt,a4paper]{article}
\usepackage[hyperref]{ranlp2017}
\usepackage{times}
\usepackage{latexsym}
\usepackage{graphicx}
\usepackage{amsmath}
\usepackage{subfigure}
\usepackage{xcolor}
\usepackage{url}
\usepackage{soul}

\aclfinalcopy 
\def\aclpaperid{15} 

\title{Neural Reranking for Named Entity Recognition}

\author{Jie Yang \and Yue Zhang \and Fei Dong\\ 
  Singapore University of Technology and Design \\
  {\tt \{jie\_yang, fei\_dong\}@mymail.sutd.edu.sg} \\
  {\tt yue\_zhang@sutd.edu.sg} \\}

\date{}

\begin{document}
\maketitle
\begin{abstract}
  We propose a neural reranking system for named entity recognition (NER). The basic idea is to leverage recurrent neural network models to learn sentence-level patterns that involve named entity mentions. In particular, given an output sentence produced by a baseline NER model, we replace all entity mentions, such as \textit{Barack Obama}, into their entity types, such as \textit{PER}. The resulting sentence patterns contain direct output information, yet is less sparse without specific named entities. For example, ``PER was born in LOC'' can be such a pattern. LSTM and CNN structures are utilised for learning deep representations of such sentences for reranking. Results show that our system can significantly improve the NER accuracies over two different baselines, giving the best reported results on a standard benchmark.
\end{abstract}

\section{Introduction}

Shown in Figure \ref{fig:nerdemo}, named entity recognition aims to detect the entity mentions in a sentence and classify each entity mention into one out of a given set of categories. NER is typically solved as a sequence labeling problem. Traditional NER systems use Hidden Markov Models (HMM) \cite{zhou2002named} and Conditional Random Fields (CRF) \cite{lafferty2001conditional} with manually defined discrete features. External resources such as gazetteers and human defined complex global features are also incorporated to improve system performance \cite{ratinov2009design,che2013named}. Recently, deep neural network models have shown the ability of learning more abstract features compared with traditional statistical models with indicator features for NER \cite{zhang2015neural}.

Recurrent Neural Network (RNN), in particular Long Short-Term Memory (LSTM) \cite{hochreiter1997long}, shows the ability to automatically capture history information over input sequences, which makes LSTM a proper automatic feature extractor for sequence labeling tasks. Different methods have been proposed by stacking CRF over LSTM in NER task \cite{chiu2015named,huang2015bidirectional,lample2016neural,ma2016end}. In addition, it is possible to combine discrete and neural features for enriched information, which helps improve sequence labeling preformance \cite{zhang2016libn3l}.

\begin{figure}[!t] 
  \resizebox{\columnwidth}{!}{%
  \begin{tabular}{c}
  \hline
  \colorbox{red!20}{\textbf{[Barack Obama]}}$_{PER}$ was born in \colorbox{blue!20}{\textbf{[hawaii]}}$_{LOC}$ .\\
  \hline
  Rare \colorbox{red!20}{\textbf{[Hendrix]}}$_{PER}$ song draft sells for almost \$ 17,000 .\\
  \hline
  \colorbox{green!20}{\textbf{[Volkswagen AG]}}$_{ORG}$ won 77,719 registrations .\\
  \hline
  \colorbox{blue!20}{\textbf{[Burundi]}}$_{LOC}$ disqualification from \colorbox{yellow!50}{\textbf{[African Cup]}}$_{MISC}$ confirmed .\\
  \hline
  The bank is a division of \colorbox{green!20}{\textbf{[First Union Corp]}}$_{ORG}$ . \\
  \hline
  \end{tabular}
  }
  \caption{Named Entity Recognition.}
  \label{fig:nerdemo}
\end{figure}

\begin{figure*}[htp] 
  \centering 
  \includegraphics[width=6.0in]{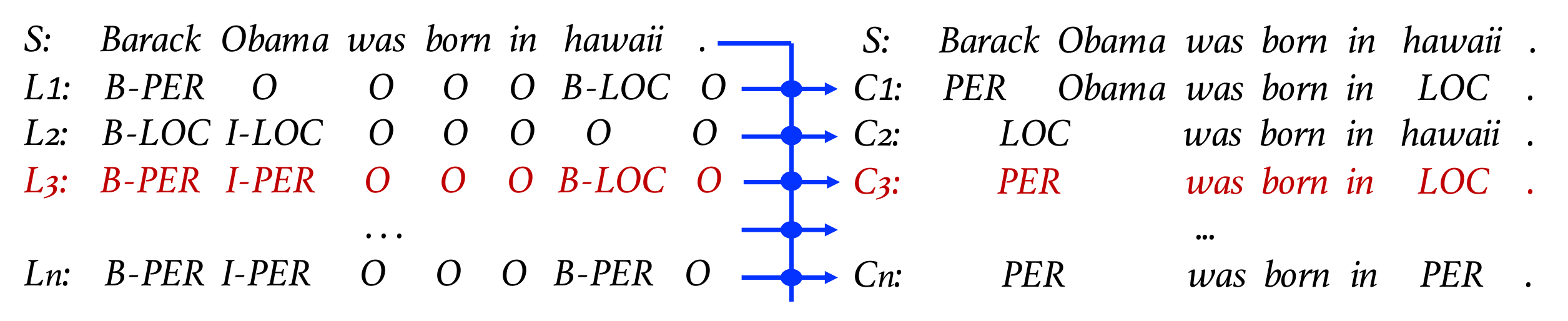}
  \caption{Example of generating collapsed sentence patterns from baseline NER output label sequences.}
  \label{fig:collapse}
\end{figure*}

Reranking is a framework to improve system performance by utilizing more abstract features. A reranking system can take full advantage of global features, which are intractable in baseline sequence labelling systems that use exact decoding. The reranking method has been used in many NLP tasks, such as parsing \cite{collins2005discriminative}, QAs \cite{chen2006reranking} and machine translation \cite{wang2007reranking,shen2004discriminative}.

Some work has adopted the reranking strategy for NER. \newcite{collins2002ranking} tried both a boosting algorithm and a voted perceptron algorithm as reranking models on named-entity boundaries (without classification of entities). \newcite{nguyen2010kernel} applied Support Vector Machine (SVM) with kernels to reranking model, obtaining significant improvements in F-measure on CoNLL 2003 datasets. \newcite{yoshida2007reranking} used a simple log-linear rerank model on a biomedical NER task, also obtaining slight improvemts. All the above methods use sparse manual features. To the best of our knowledge, there has been no neural reranking model for NER task.

In this paper, we propose a simple neural reranking model for NER. The model learns sentence patterns that involve output named entities automatically, using neural network. Take the sentence \textit{\ul{``Barack Obama was born in hawaii .''}} as an example, Figure \ref{fig:collapse} illustrates several candidate sentence patterns such as \textit{\ul{``PER was born in LOC .''}} ($C_3$) and \textit{\ul{``LOC was born in hawaii .''}} ($C_2$), where \textit{PER} represents entity type \textit{persons} and \textit{LOC} means \textit{locations}. It is obvious that $C_3$ is a much more reasonable sentence pattern compared to $C_2$. To generate the sentence patterns above, we replace predicted entities in candidate sequences with their entity type names. This can effectively reduce the sparsity of candidate sequences, as each entity type contains open vocabulary names (e.g. \textit{PER} can be \textit{Donald Trump}, \textit{Hillary Clinton} etc.), which can bring noise when learning the sentence patterns. In addition, since the learned sentence patterns are global over output structures, it is difficult for baseline sequence labeling systems to capture such patterns.

We develop a neural reranking model which captures candidate pattern features using LSTM and auxilliary neural structures, including Convolution Neural Network (CNN) \cite{kim2014convolutional,kalchbrenner2014convolutional} and character based neural features. The learned global sentence pattern representations are then used as features for scoring by the reranker. Results over a state-of-the-art discrete baseline using CRF and a state-of-the-art neural baseline using LSTM-CRF show significant improvements. On CoNLL 2003 test data, our model achieves the best reported result.

Our main contributions include (a) leveraging global sentence patterns that involve entity type information for NER raranking, (b) exploiting auxilliary neural features to enrich basic LSTM sequence representation and (c) achieving the best F1 result on CoNLL 2003 data. The source codes of this paper are released under GPL at \url{https://github.com/jiesutd/RerankNER}.

\begin{table}[t]
\begin{center}
\resizebox{\columnwidth}{!}{%
\begin{tabular}{|l|l|}
\hline 
\bf Description& \bf Feature Template  \\ 
\hline
word grams & $w_i,w_iw_{i+1}$\\
shape, capital & $Sh(w_i),Ca(w_i)$\\
capital + word & $Ca(w_i)w_i$\\
connect word & $Co(w_i)$\\
capital + connect & $Ca(w_i)Co(w_i)$\\
cluster grams & $Cl(w_i),Cl(w_iw_{i+1})$\\
prefix, suffix & $Pr(w_i), Su(w_i)$\\
POS grams & $P(w_i,w_iw_{i+1},w_{i-1}w_1w_{i+1})$\\
POS + word & $P(w_0)w_0$\\
\hline
\end{tabular}
}
\end{center}
\caption{Features of discrete \textit{CRF} for NER, $i\in\{-1,0 \}$.}
\label{tab:features}
\end{table}

\section{Baselines}
Formally, given a sentence $S$ with $t$ words: $S=\{w_1,w_2,...,w_t\}$, the task of NER is to find out all the named entity mentions from $S$. The dominate approach takes the task as a sequence labelling problem, where the goal is to generate a label sequence $L = \{l_1, l_2, ..., l_t\}$, where $l_i = p_i e_i$. Here $p_i$ is an entity label, $p_i \in \{B, I, O\}$, where $B$ indicates the beginning of an entity mention, $I$ denotes a non-beginning word of a named entity mention and $O$ denotes a non-named-entity word \footnote{When $p_i=O$, $e_i$ equals to NULL.}. $e_i$ indicates the entity type. In the CoNLL dataset that we use for our experiments,
 $e_i \in\{ \it PER, ORG, LOC, MISC\} $, where ``$PER$'' indicates a \textit{person} name; ``$LOC$'', ``$ORG$'', ``$MISC$'' represent \textit{location}, \textit{organization} and \textit{miscellaneous}, respectively.

We choose two baseline systems, one using dicrete CRF with handcrafted features and one using neural CRF model with bidirectional LSTM structure, both baselines giving the state-of-the-art accuracies among their respective category of models.

\subsection{Discrete CRF}
We choose a basic discrete CRF model as our baseline tagger. As shown in Figure \ref{fig:discretecrf}, discrete word features are first extracted as binary vectors (black and white circles) and then fed into a CRF layer. Taking those discrete features as input, the CRF layer can give \textit{n-best} predicted sequences as well as their probabilities. Table \ref{tab:features} shows the discrete features that we used, which follow the definition of \cite{Jie2016combining}. Here \textit{shape} means whether characters in word are belonging to number, English character or not. \textit{capital} is the indication if word starts with \textit{upper-case} English character, \textit{connect words} include  five types: ``of'', ``and'', ``for'', ``-'' and other. Prefix and suffix include the 4-level prefixes and suffixes of each words.

\subsection{Neural CRF}
\begin{figure}[!t] 
  \centering 
  \subfigure[Discrete CRF baseline.]{ 
    \label{fig:discretecrf} 
    \includegraphics[width=2.8in]{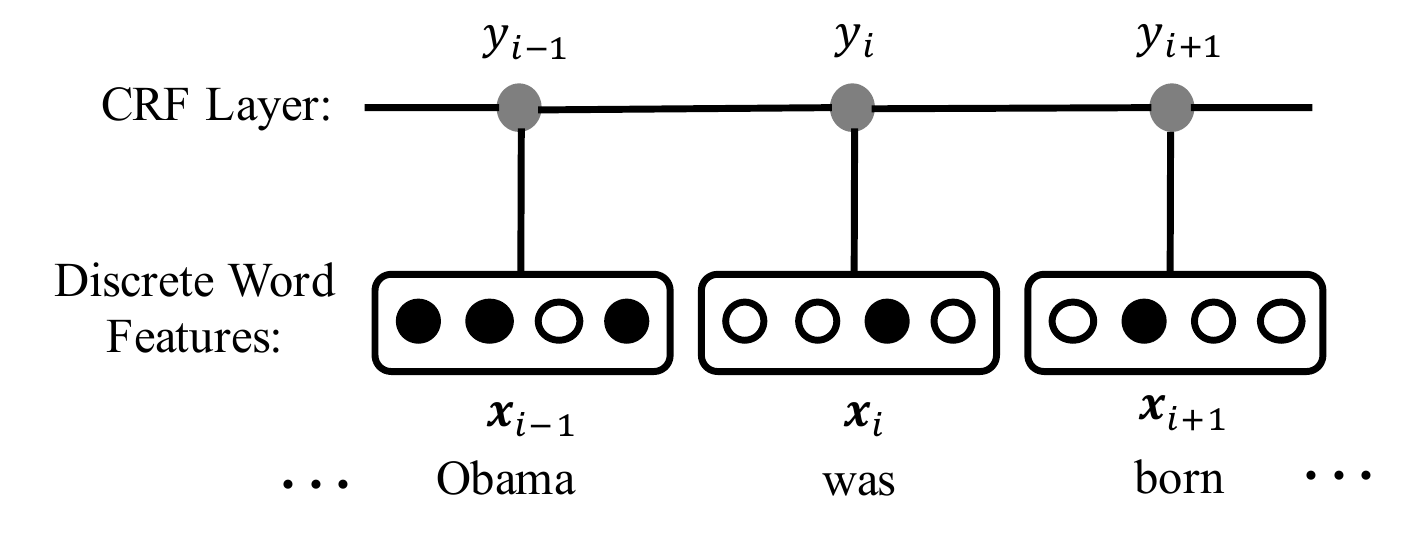}} 
  \subfigure[Neural CRF baseline.]{ 
    \label{fig:neuralcrf} 
    \includegraphics[width=2.8in]{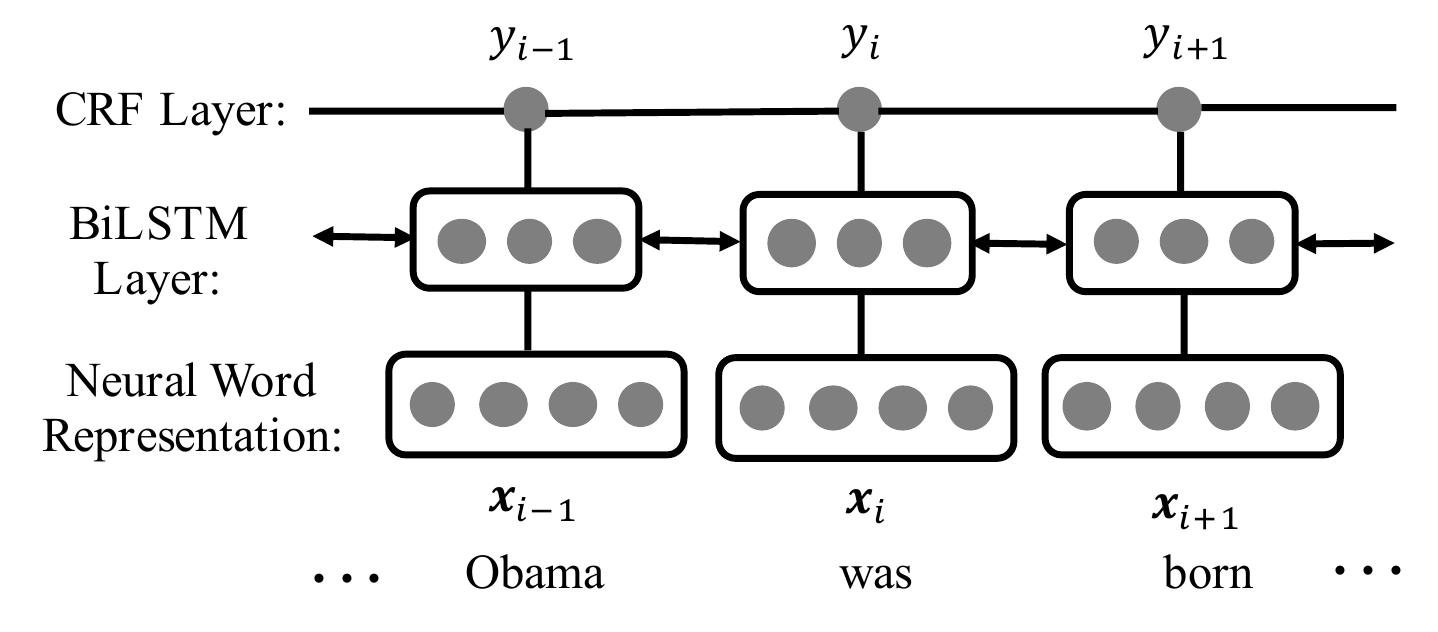}} 
  \caption{Baselines.} 
  \label{baselines} 
\end{figure}

A neural CRF with bidirectional LSTM structure is used as our second baseline, which is shown in Figure \ref{fig:neuralcrf}.  Word representations are represented with continious vectors (gray circles), which are fed into a bidirectional LSTM layer to extract neural features. A CRF layer with \textit{n-best} output is stacked on top of the LSTM layer to decode the label sequences based on the neural features. We use the neural structure of \citet{ma2016end}, where the word representation is the concatenation of word embedding and a CNN output on the character sequence of the word.

\section{Reranking Algorithms}
\subsection{Collapsed Sentence Representation}\label{rule}
Given the \textit{n-best} output label sequences of a baseline system $\{L_1,L_2,...,L_i,...,L_n\}$, where $L_i = \{l_{i1},l_{i2},...,l_{it}\}$, we learn a reranking score $s(L_i)$ for $L_i$ by firsting converting $L_i$ into a sequence pattern $C_i$, and then learning a representation $h(C_i)$ as its dense representation. To convert candidate sequence $L_i$ to collapsed sequence $C_i$. We use the following rules to convert each label sequence $L_i$ into a collapsed sentence pattern $C_i$. 

If the $L_i$ include entity labels (e.g. $l_{i1}$ =\textit{B-PER}, $l_{i2}$=\textit{I-PER}), then the entity labels are replaced with the corresponding entity type name (e.g. $\{l_{i1},l_{i2}\}$ $\rightarrow$ \textit{PER},  $C_{i1}$ = \textit{PER}), else labels are replaced by its corresponding words ($C_{ix} = w_x$). In the example shown in Figure \ref{fig:collapse}, $S=$ \textit{\{Barack Obama was born in hawaii .\}} and $L_3=$ \textit{\{B-PER I-PER O O O B-LOC O\}}. The corresponding collapsed sequence is $C_3=$ \textit{\{PER was born in LOC .\}}, \textit{Barack Obama} and \textit{hawaii} are regarded as entities and hence are replaced by the entity names, i.e. \textit{PER} and \textit{LOC}, respectively. 


\subsection{Neural Features}
Given a collapsed sentence representation $C_i$, we use neural network to learn its overall representation vector $h(C_i)$, which is used for the scoring of $C_i$. 

\textit{\textbf{Word Representation:}} We use \textit{SENNA} \cite{collobert2011natural} embedding to initialize the word embedding of our reranking system. For out of vocabulary words , embeddings are randomly initialized within $(-\sqrt{\frac{3.0}{wordDim}},\sqrt{\frac{3.0}{wordDim}})$, where $wordDim$ is the word dimension size \cite{ma2016end}. 

\begin{figure}[!t] 
  \centering 
  \subfigure[CNN character sequence representation for word.]{ 
    \label{fig:cnnchar} 
    \includegraphics[width=2.8in]{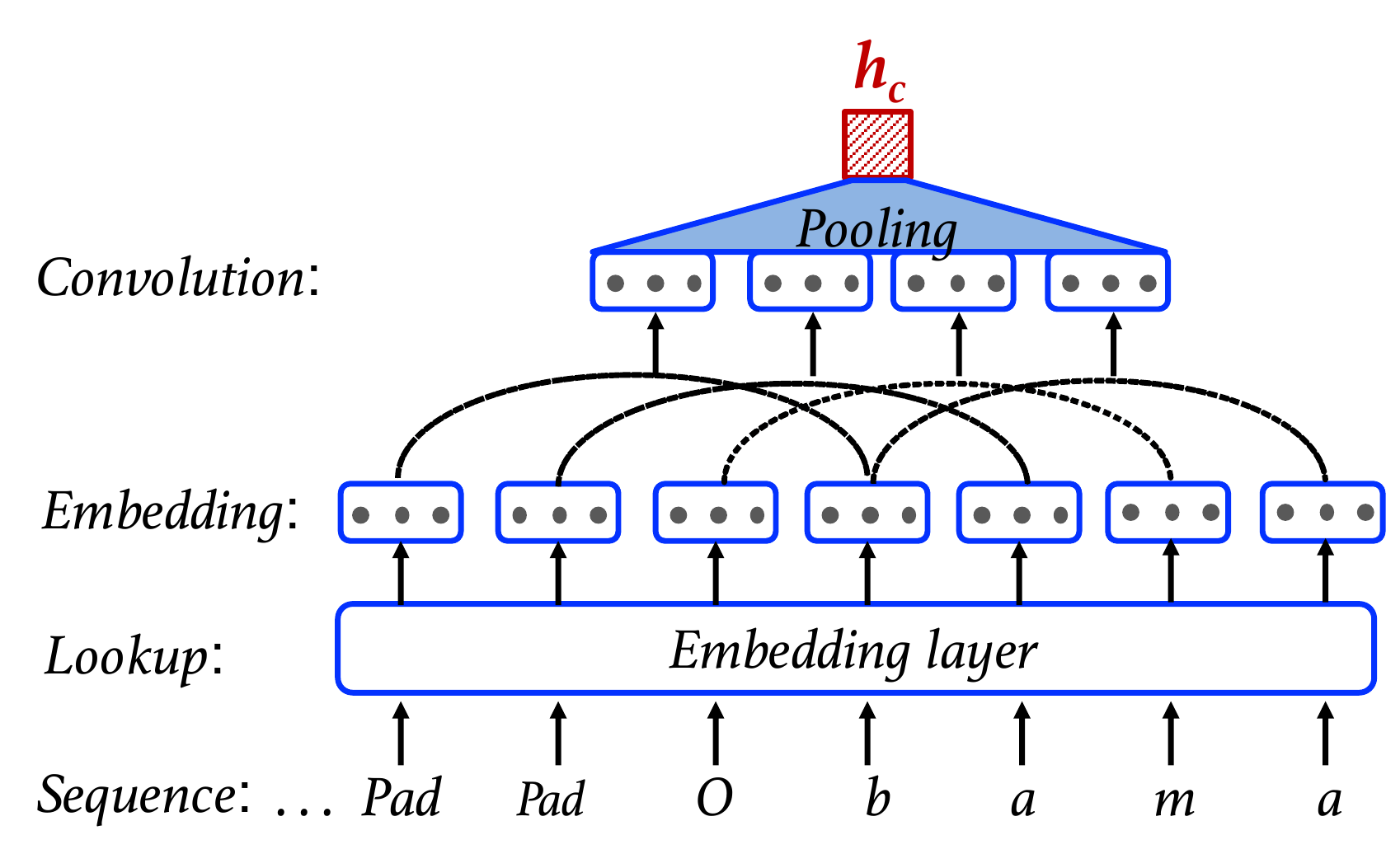}} 
  \subfigure[LSTM word sequence representation for sentence.]{ 
    \label{fig:lstmword} 
    \includegraphics[width=2.8in]{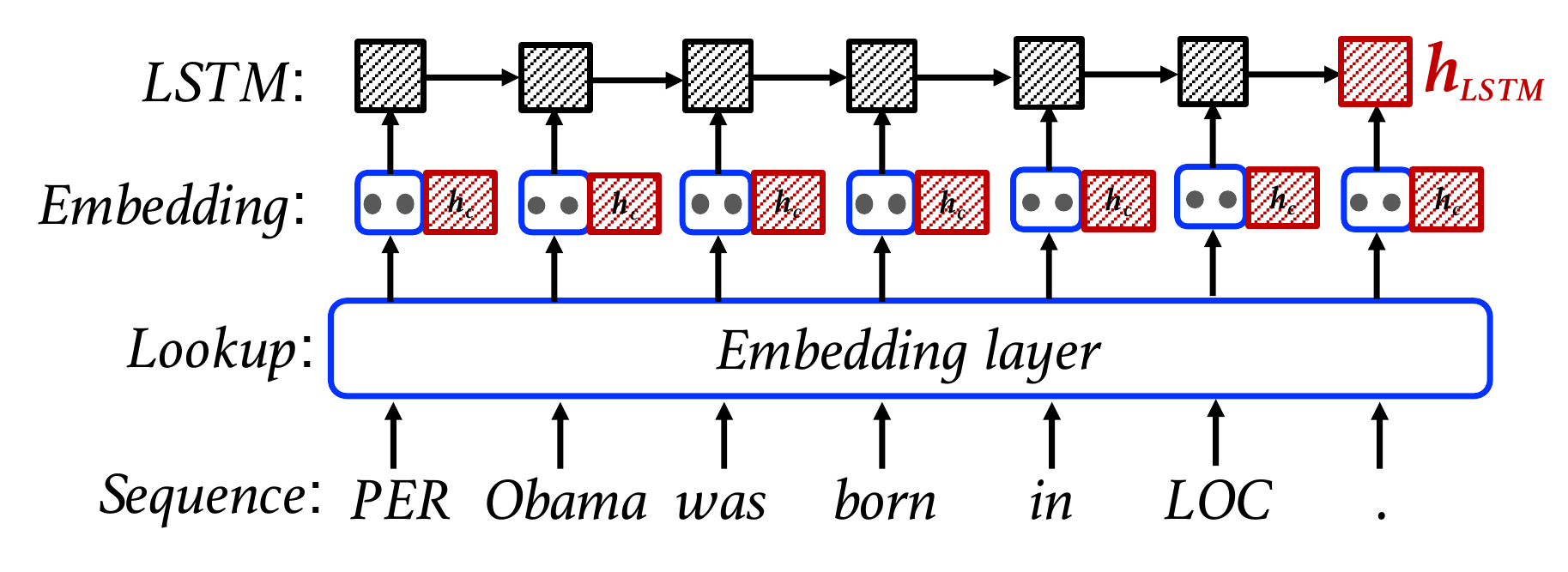}} 
  \caption{Representation.} 
  \label{representation} 
\end{figure}

Character features are proved useful in capturing morphological features, such as word similarity and dealing with the out-of-vocabulary problem \cite{ling2015finding}. As shown in Figure \ref{fig:cnnchar}, we follow \citet{ma2016end} by utilizing CNN to extract character-level representation \footnote{Characters are padded into a fixed length by using a special token \textit{Pad}.}. Input character sequences are firstly passed through the embedding layer to lookup the character embeddings. To extract local features, a \textit{convolution layer} with a fixed window-size is applied on top of the embedding layer. Then we use a \textit{max-pooling} layer to map varying length vectors into a fixed size output vector. Finally, word representation is the concatenation of character CNN output vectors and word embeddings.

\textit{\textbf{LSTM features:}} We choose a word-based LSTM as the main network, using it for capturing global sentence pattern information.
For input sequence vectors $\{x_1,x_2,...,x_t\}$, our LSTM model is implemented as follows:
\begin{equation*}
\begin{aligned}
h_t& \:=\:tanh(M_t)\odot o_t\\
i_t&\:=\:\sigma(W_1h_{t-1}+W_2x_t+\mu_1 \odot M_{t-1}+b_1)\\
f_t&\:=\:\sigma(W_3h_{t-1}+W_4x_t+\mu_2 \odot M_{t-1}+b_2)\\
\widetilde{M_i}&\:=\:tanh(W_5y_{t-1}+W_6x_i+b_3)\\
M_t&\:=\:i_i\odot\widetilde{M_i}+f_i\odot M_{t-1}\\
o_t&\:=\:\sigma(W_7h_{t-1}+W_8x_t +b_4),\\
\end{aligned}
\end{equation*}
where $\odot$ is the element-wise multiply operator, $\sigma$ is the sigmoid function, and $\{W,b,\mu\} \in \Theta$ are parameters. $i_t, f_t, M_t$ and $o_t$ are the \textit{input gate, forget gate, memory cell} and \textit{output gate}, respectively. $h_t$ is the hidden vector at step $t$ in the input sentence.
As shown in Figure \ref{fig:lstmword}, word representations are the concatenation of word embeddings and character CNN output (red block). We choose the hidden vector in last word $h_{LSTM}$ as the representation of the input sequence. 

\textit{\textbf{CNN features:}} We introduce CNN to capture local features of the candidate sequences. It consists of a \textit{filter} 
$W \in R^{h\times k}$ which operates on a context of $k$ words
to produce local order features. Max pooling layer is employed over the convolutional layer to 
extract the most salient features. Assume $u_j$ is the concatenation of word representations in Eq. (1) centralized in the embedding $z_j$ in a given sequence $u_1, u_2, ..., u_L$,
CNN applies a matrix-vector operation to each window of size $k$ successive window along the sequence in Eq. (2).
\begin{align}
u_j &= (z_{j-(k-1)/2}, ..., z_{j+(k-1)/2}) \\
r_i &= max_{1<j<L} {(W u_j  + b)_i}, i=1, ..., d ,
\end{align}
where $z_j$ is the $j$-th word embedding in the given sequence,
$d$ is the output dimension of the CNN. $h=[r_1, ..., r_i,..., r_d]$ is the fixed-size
feature representation for the sequence after pooling.

The \textit{CNN} representation structure is similar to Figure \ref{fig:cnnchar} but its input is word representations rather than character embeddings. We define the CNN features of word sequence as $h_{CNN}$.

\subsection{Score Calculation}
After the LSTM and CNN features of collapsed sequence $C_i$ are extracted, we concatenate them together and feed the result into a \textit{softmax} layer.

\begin{equation}
\begin{aligned}
h(C_i) & \:=\:h_{LSTM}\oplus h_{CNN}\\
s(C_i) & \:= \: \sigma(W h(C_i)  + b) ,
\end{aligned}
\end{equation}
where $\oplus$ represents the concatenating operation, $h(C_i)$ is the final representation of collapsed sequence $C_i$ and $s(C_i)$ is the output score of $C_i$.

\subsection{Decoding}
We use a mixture reranking strategy during decoding. Denote the candidate label sequence set on sentence $S$ as $C(S) = \{C_1,C_2,...,C_n\}$ . We take advantage of both the reranker prediction score and the baseline tagger's output probability, using the score
\begin{equation}
\label{equ:mixture}
\hat{y_i} = \arg\max_{C_i\in C(S)} (\alpha s(C_i) + (1-\alpha)p(L_i)),
\end{equation}
where $\alpha \in [0,1]$ is an interpolation weight, which is a hyperparameter tuned on the development set. $p(L_i)$ is the probability of label sequence $L_i$ in the baseline tagger.

\begin{table}[t]
\begin{center}
\resizebox{\columnwidth}{!}{%
\begin{tabular}{|l|l|l|l|}
\hline \bf Parameter& \bf Value  &\bf Parameter& \bf Value\\ \hline
\textit{n-best}& 10 &peepholes& no\\
wordDim& 50 &charDim& 50\\
LSTM hidden & 100  &dropout & 0.2\\
charCNN filter& 50  &batch size& 128\\
wordCNN filter & 100  &$\lambda$& 0.001 \\
charCNN length & 3 &\textit{Adam} $\beta_1$ & 0.1\\
wordCNN length&3 &\textit{Adam} $\beta_2$ & 0.999  \\
learning rate & 0.001 &\textit{Adam} $\epsilon$ & 1e-8 \\
\hline
\end{tabular}
}
\end{center}
\caption{Hyperparameters of reranker.}
\label{tab:hyperparameter}
\end{table}

\subsection{Training}
For each training triplet $\{S,L_i, C_i\}$, given the golden sequence $L_{golden}$, we calculate the tag accuracy $y_i \in [0,1]$  of each candidate sequence based on $L_i$ and $L_{golden}$. The same decoding process is applied to each collapsed sequence $(C_i, y_i)$. We use a logistic regression model with mean square error (MSE) as the loss function, with a $l_2$-regulation term \footnote{We also tried \textit{max-margin} criterion like \cite{zhu2015re}, while the results are similar with regression model.} :
\begin{equation}
J(\Theta) \:=\: \frac{1}{|\mathcal{D}|}\sum_{(C_i, y_i)\in \mathcal{D}}(y_i-s(C_i))^2 + \frac{\lambda}{2}||\Theta||_2^2
\end{equation}
where $\Theta$ are all the parameters to be trained, $\mathcal{D}$ is the training set and $\lambda$ is the regulation factor. 

\textit{Adam} \cite{kingma2014adam} is used to update model parameters.

\section{Experiments}
\subsection{Settings}
We use \textit{CRF++} \footnote{https://taku910.github.io/crfpp/} as our discrete baseline CRF implementation and default parameters are used. For neural baseline, we follow the same structure and settings of the state-of-the-art system \cite{ma2016end}. When building the neural reranking system, \textit{SENNA} embedding with 50 dimensions is used to initialize word embeddings. Hyperparameters of reranking system are listed in Table \ref{tab:hyperparameter}.

As we use the mixture strategy in Eq. (\ref{equ:mixture}) during decoding, we search the ideal interpolation weight $\alpha$ within $[0,1]$ in a step of \textit{0.005} based on the preformance under the development set.

\subsection{Reranking Data}
All of our experiments are evaluated on the standard CoNLL 2003 English dataset \cite{tjong2003introduction}, which is a collection of Reuters newswire articles. The CoNLL 2003 English dataset includes 14,987 training sentences, 3,466 development sentences and 3,684 test sentences, annotated into 4 entity types, i.e. \textit{persons(PER)}, \textit{locations(LOC)}, \textit{organizations(ORG)} and \textit{miscellaneous(MISC)}.

To construct the reranking training data, we conduct five-fold \textcolor{black}{jackknifing}, spliting the training set into 5 equal parts. In each case, the baseline tagger trains the model with 4/5 of the data and decode the remaining 1/5 to generate \textit{n-best} candidate label sequences. For the reranking development and test data, the full training set is used to encode baseline tagger and decode development/test sentences with \textit{n-best} output. All the \textit{n-best} candidate sequences are converted into collapsed sequences following Section \ref{rule}.


\begin{figure}[t] 
  \centering 
  \includegraphics[width=2.5in]{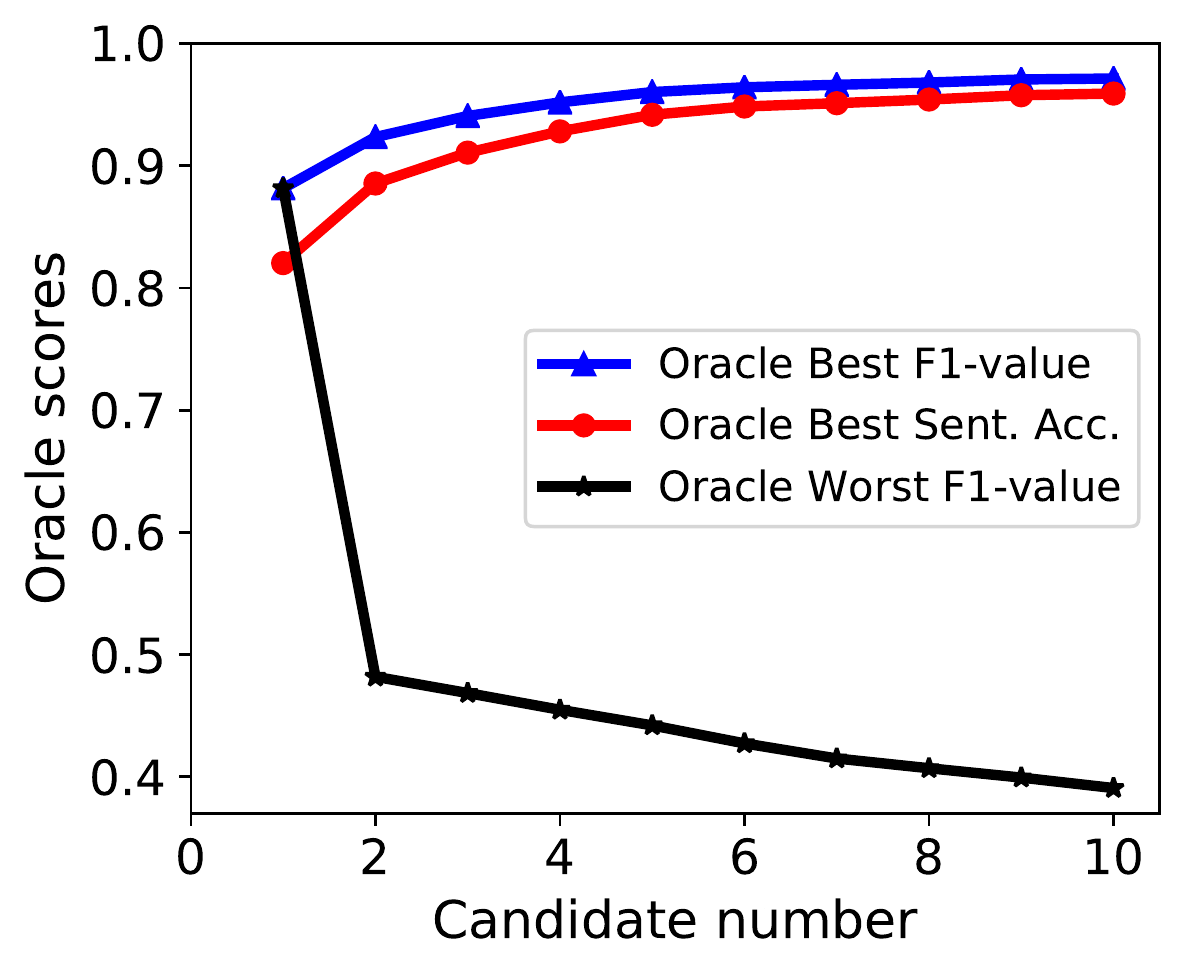}
  \caption{Oracle scores in baseline outputs.}
  \label{fig:baseline}
\end{figure}

\subsection{Baseline Oracle Results}
The discrete baseline achieves 92.13\% of F1-measure in development set and 88.15\% in test set. Our neural baseline gives 94.58\% and 91.25\% on development and test data, respectively. The discrete baseline for example. Figure \ref{fig:baseline} shows different oracle scores varying with \textit{n-best} in discrete baseline. The oracle best is obtained by always chooseing the best sequence in the \textit{n-best} candidates, and \textit{vice versa} for the oracle worst. The orcale best sentence  accuracy (OBA) \footnote{Notice this is different with accuracy which represents the correct rate of tags, OBA represents the correct rate in sentence level.} represents the accuracy of the sequence choice within the \textit{n-best} candidates under oracle best assumption, and the orcale best F1-value (OBF) is the corresponding F1-value similarly. Orcale worst F1-value (OWF) is the F1-value under the worst choice situation. 

As the figure shows, the larger $n$ is, the better is the OBA, which means that a potentially better reranking result is possible. On the other hand, the OWF also drops, which means that the reranking task is more difficult. In our experiments, \textit{n-best} is set as 10, the orcale best F1-value of test set achieves 97.13\% (+8.98\%) while its orcale worst F1-value drops 49.07\% to 39.08\%. 

\begin{figure}[t] 
  \centering 
  \includegraphics[width=3in]{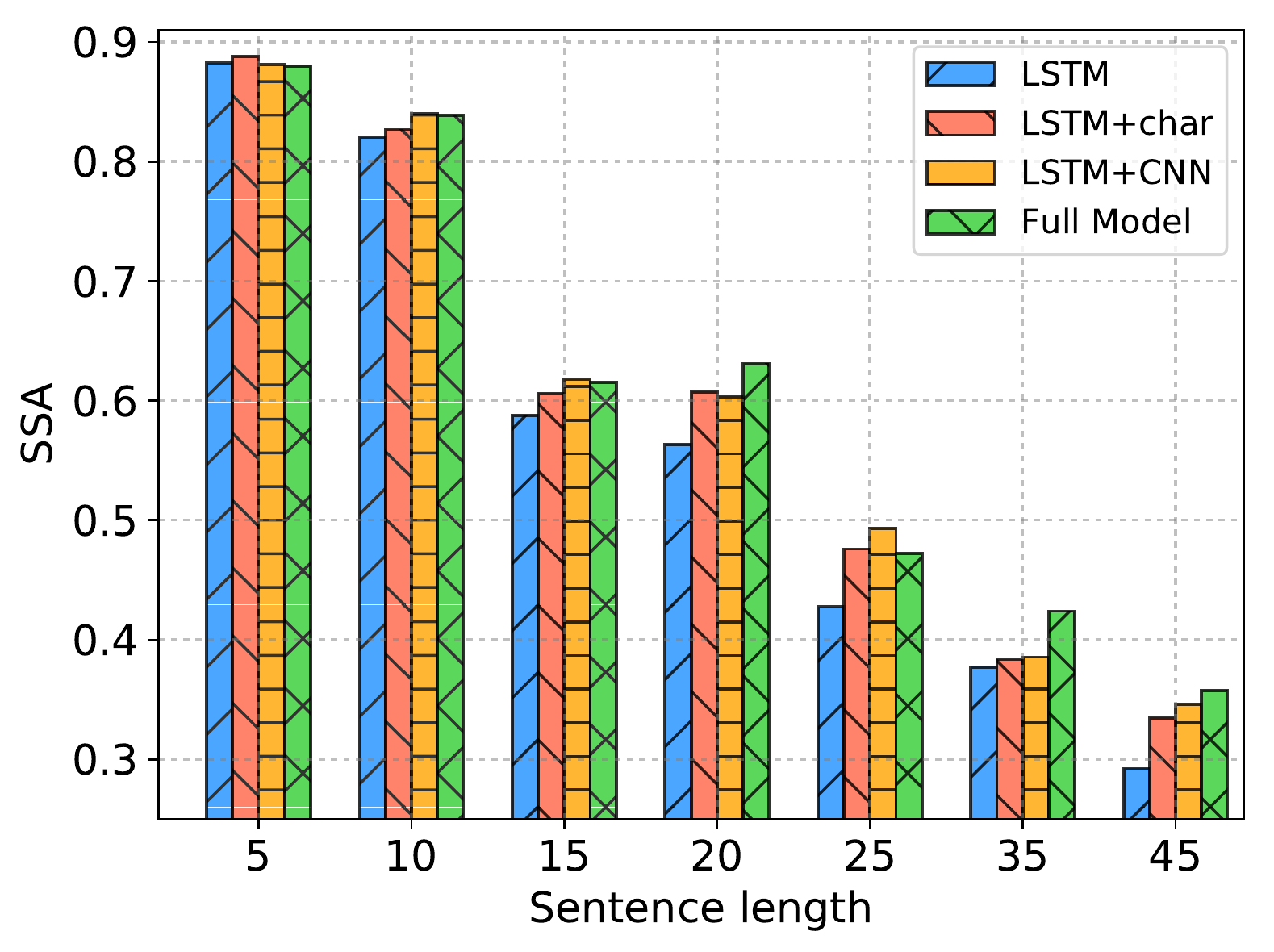}
  \caption{SSA with sentence length.}
  \label{fig:sentAcc}
\end{figure}

\subsection{Influence of Sentence Length}
We perform development experiments to evaluate model performance on various sentence lengths. Figure \ref{fig:sentAcc} shows results by reranking the discrete baseline. Here sentence select accuracy (SSA) is calculated using the corrected number of sentences divided by the total number of sentences.
The \textit{x-axis} is the sentence length range (e.g. 10 means sentence length range from 5 to 10), while the \textit{y-axis} corresponds to SSA within 10-best candidates before the mixture strategy (without mixing baseline output probability). 

As shown in the Figure \ref{fig:sentAcc}, the accuracies of all model settings drop as the size of the sentence increases, which demonstrates that longer sentences are more challenging to our neural rerankers as they are to the baseline models. \textcolor{black}{This is because for longer sentences, candidate collapsed sequences have higher overlapped proportion and hence are hard to be distinguished by reranker.}  Both character information and CNN local features are useful for enhancing the SSA over a LSTM-only baseline. With the integration of character information and CNN features, our full model reranker can improve its performance on all sentence length ranges, especially for long sentences.

\begin{figure}[t] 
  \centering 
  \includegraphics[width=3in]{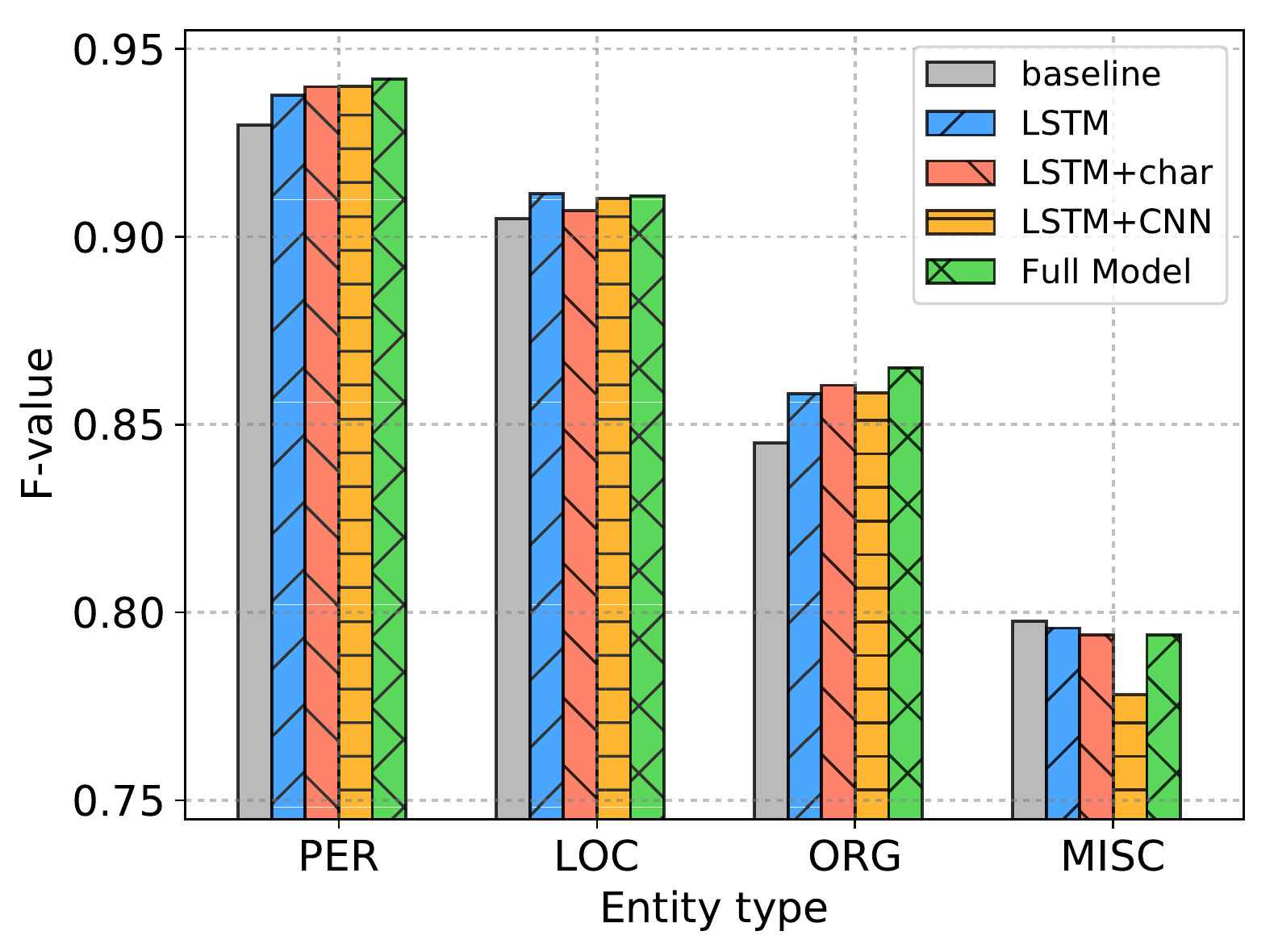}
  \caption{F1-value comparision by entity types.}
  \label{fig:entityAnalysis}
\end{figure}

\subsection{Influence of Entity Type}
Figure \ref{fig:entityAnalysis} shows the comparision of models on different entity types. Compared with the baseline, entities with type of \textit{PER} and \textit{ORG} receive the most improvements, showing that sentence patterns are useful for those types. \textcolor{black}{While the performance on entities with type of \textit{MISC} decreases slightly, since the \textit{MISC} includes various entity types which bring noise on learning sentence patterns. We believe that our model will benifit more from the NER corpus with fine-grained entity type.} Table \ref{tab:finalresult} shows the F1-value and SSA (after mixing baseline output probability) of our reranker on test data with different neural features on the discrete baseline. The word based LSTM reranker achieves the F1-value of 88.75\%, with 0.6\% absolute improvement over the baseline tagger. Cooperating with CNN features on word only does not make much improvement, while character CNN features are more effective (+0.78\%). However, the full combination of character representation and word CNN features improves the F-value to 89.25\% (+1.10\%) with the significance level of $p<0.05$ with \textit{t-test}. The trend of SSA is the same as the F1-value, the accuracy is improved from the baseline 83.31\% to 85.12\% using the full model reranker, with an absolute improvement of 1.82\%.

\begin{table}[t]
\begin{center}
\resizebox{\columnwidth}{!}{%
\begin{tabular}{|l|l|l|l|l|}
\hline \bf Model (\%)& \bf F1 & \bf$\Delta$F1 & \bf SSA & \bf$\Delta$SSA\\ \hline
\textit{Baseline}& 88.15 &0 &83.31  & 0\\
\textit{LSTM} & 88.75 &0.60 &84.41 & 1.10\\
\textit{LSTM+CNN} & 88.79 &0.64 &84.63&1.32\\
\textit{LSTM+char} &  88.93 &0.78 &84.69&1.38\\
\textit{Full model} & \bf 89.25 &1.10 & \bf 85.12 &1.82 \\
\hline
\end{tabular}
}
\end{center}
\caption{F1-value and SSA on test set.}
\label{tab:finalresult}
\end{table}

\subsection{Effectiveness of Reranking}
Table \ref{tab:bestresult} shows our rerank results on two baselines and the comparison with state-of-the-art systems. Our reranker on discrete baseline compares favourably to the best discrete models, including the use of external corpus \cite{kazama2007exploiting,suzuki2008semi}. It also outperforms \citet{nguyen2010kernel} which builds a discrete reranking model by utilizing SVM with kernels. \citet{ratinov2009design}* achieves 90.57\% in discrete model by combining global features and abundant external lexicons, while its performance drops to 88.55\% when removing the global features \cite{ratinov2009design}. \citet{luo2015joint} gives the best discrete result (91.20\%) by jointing NER with disambiguation task together. 

\begin{table}[t]
\begin{center}
\begin{tabular}{|c|c|}
\hline 
\bf Discrete Model (\%)& \bf F1 \\
\hline
\citet{kazama2007exploiting}&88.02\\ 
\citet{suzuki2008semi}&89.92\\ 
\citet{nguyen2010kernel}&88.16 \\ 
\citet{ratinov2009design}&88.55 \\
\citet{ratinov2009design}*&90.57\\  
\citet{luo2015joint} & 91.20 \\
\hline
Discrete baseline&  88.13 \\ 
Our reranker&  89.25 \\
\hline
\hline
\bf Neural Model (\%)& \bf F1 \\ 
\hline
\citet{collobert2011natural}&89.59 \\
\citet{passos2014lexicon} & 90.90 \\
\citet{huang2015bidirectional}& 90.10 \\
\citet{chiu2015named} & 90.77 \\
\citet{lample2016neural} & 90.94\\
\citet{ma2016end} & 91.21\\
\hline
Neural baseline&  91.25\\
Our reranker&  \bf91.62\\
\hline
\end{tabular}
\end{center}
\caption{Comparison of state-of-the-art systems.}
\label{tab:bestresult}
\end{table}

\citet{collobert2011natural} builds a first neural NER model with comparable performance to discrete models on CoNLL 2003 corpus. Most state-of-the-art neural NER models utilize bidirectional LSTM with a CRF layer \cite{huang2015bidirectional}. \citet{lample2016neural} and \citet{ma2016end} concatenate character representation with word embedding and \citet{chiu2015named} even merge lexicon features into word representation. \citet{passos2014lexicon} obtain a 90.90\% by combining discrete features and neural word embeddings in a CRF model. Our neural baseline, which takes the same features as \citet{ma2016end}, achieves 91.25\% in F-value. Our reranker on this baseline outperforms all the previous models with the F-value of 91.62\%, which is the best reported F-score on CoNLL 2003.


\subsection{Examples}
Figure \ref{fig:resultexample} gives some example outputs on the development dataset for which discrete baseline gives incorrect outputs yet the reranker corrects the mistake. Our reranker learns better sentence patterns by correcting both named entity boundary errors and named entity type errors. 

In the first case, example 1 shows that \textit{``U.N. Ambassador Albright''} in sentence \textit{\ul{``U.N. Ambassador Albright arrives in Chile .''}} is incorrectly tagged as a \textit{organization} by the baseline and the entity boundary is incorrect either. By building the collapsed sentences as the input of our reranker, entities such as \textit{``U.N. Ambassador Albright''} are replaced as a single entity name \textit{``ORG''}. Our reranking model learns that \textit{\ul{``... PER arrives in LOC ...''}} is more possible compared to \textit{\ul{``... ORG arrives in LOC ...''}}, thereby the candidate with the reasonable entity boundary and type is picked by our reranker.

For the second case, the entity type of \textit{``EL SALVADOR''} in example 3 \textit{\ul{``SOCCER - U.S. BEAT EL SALVADOR 3-1 .''}} is incorrectly recognized as \textit{organization} by baseline. Our reranker corrects this entity type error by giving higher score to sentence pattern \textit{\ul{``... LOC BEAT LOC ...''}} rather than pattern \textit{\ul{``... LOC BEAT ORG ...''}}.

\begin{figure}[!t] 
  \resizebox{\columnwidth}{!}{%
  \begin{tabular}{|l|l|}
  \hline
  \textbf{Baseline\hfill 1}&\colorbox{green!20}{\textbf{[U.N. Ambassador Albright]}}$_{ORG}$ arrives in \colorbox{blue!20}{\textbf{[Chile]}}$_{LOC}$ .\\
  \textbf{Reranker 1}&\colorbox{green!20}{\textbf{[U.N.]}}$_{ORG}$ Ambassador \colorbox{red!20}{\textbf{[Albright]}}$_{PER}$ arrives in \colorbox{blue!20}{\textbf{[Chile]}}$_{LOC}$ .\\
  \hline
  \textbf{Baseline \hfill 2} &West \colorbox{yellow!50}{\textbf{[Indian]}}$_{MISC}$ all-rounder \colorbox{red!20}{\textbf{[Phil Simmons]}}$_{PER}$ took four ...\\
  \textbf{Reranker 2}&\colorbox{yellow!50}{\textbf{[West Indian]}}$_{MISC}$ all-rounder \colorbox{red!20}{\textbf{[Phil Simmons]}}$_{PER}$ took four ...\\
  \hline 
  \hline
  \textbf{Baseline\hfill 3}&SOCCER - \colorbox{blue!20}{\textbf{[U.S.]}}$_{LOC}$ BEAT \colorbox{green!20}{\textbf{[EL SALVADOR]}}$_{ORG}$ 3-1 .\\
  \textbf{Reranker 3}&SOCCER - \colorbox{blue!20}{\textbf{[U.S.]}}$_{LOC}$ BEAT \colorbox{blue!20}{\textbf{[EL SALVADOR]}}$_{LOC}$ 3-1 .\\
  \hline
  \textbf{Baseline\hfill  4}&... prisoners are held in \colorbox{blue!20}{\textbf{[Rangoon]}}$_{LOC}$ 's \colorbox{red!20}{\textbf{[Insein Prison]}}$_{PER}$ .\\
  \textbf{Reranker 4}&... prisoners are held in \colorbox{blue!20}{\textbf{[Rangoon]}}$_{LOC}$ 's \colorbox{blue!20}{\textbf{[Insein Prison]}}$_{LOC}$ .\\
  \hline
  \textbf{Baseline\hfill 5}&\colorbox{blue!20}{\textbf{[PAKISTAN]}}$_{LOC}$ WIN TOSS , PUT \colorbox{green!20}{\textbf{[ENGLAND]}}$_{ORG}$ INTO BAT.\\
  \textbf{Reranker 5}&\colorbox{blue!20}{\textbf{[PAKISTAN]}}$_{LOC}$ WIN TOSS , PUT \colorbox{blue!20}{\textbf{[ENGLAND]}}$_{LOC}$ INTO BAT.\\
  \hline
  \end{tabular}
  }
  \caption{Output examples. The first two examples illustrate the correction of entity boundary errors and the followings show the correction of entity type errors.}
  \label{fig:resultexample}
\end{figure}

\section{Conclusion}
We proposed a neural reranking architecture for NER by exploiting neural structure to learn sentence patterns. Given the candidate label sequences generated from a baseline tagger, we replace the predicted entity words with the corresponding entity type names to build collapsed sentences, which are used as inputs of a neural reranking model. A mixture reranking strategy is used to combine both the knowledge of the probability from the baseline tagger and the reranker score. Experiments on both discrete and neural baselines show our reranking system improves NER performance significantly, obtaining the best results on CoNLL 2003 English task .

\textcolor{black}{
One problem of current method is that all the candidates share the same non-entity words, which lead their neural representations quite similar and hard to distinguish, especially for long sentences. In the future work, we will develop the neural tree structures based on entity position which can enlarge the difference between candidate sequences. Intuitively, we believe the entities contribute more than non-entity when modeling the sequence vector, \textit{attention} model \cite{bahdanau2014neural} may help collect more information from the intermediate vector of sentences. Besides, Using semi-supervised methods to construct a bigger training data can help reranker learn more sentence patterns. Moreover, we also want to bring in an auxilliary classifier of predicting the probability of the replaced words being a real entity, this inside-entity information may be an important compensation for the outside-entity sentence patterns.
}


\section*{Acknowledgments}
\textcolor{black}{
We thank the anonymous reviewers for their insightful comments. Yue Zhang is the corresponding author.
}

\bibliography{ranlp2017}
\bibliographystyle{acl_natbib}
\end{document}